\documentclass[10pt,leqno]{amsart}
\usepackage{graphicx}
\usepackage{float}       
\usepackage{tabularx}    
\usepackage{booktabs}
\usepackage{array}
\usepackage{multirow}
\usepackage{indentfirst}
\usepackage{csquotes}
\usepackage{amssymb}
\usepackage{amsthm}
\usepackage{amsmath}

\newcolumntype{L}{>{\raggedright\arraybackslash}X}
\newcolumntype{C}{>{\centering\arraybackslash}X}
\newcolumntype{R}{>{\raggedleft\arraybackslash}X}

\usepackage{xcolor}
\usepackage{paralist}
\usepackage{hyperref}
\usepackage{fancyhdr}
\usepackage{etoolbox}

\usepackage{graphicx}
\graphicspath{{figures/}}

\hypersetup{
    colorlinks=true,
    linkcolor=black,
    citecolor=black,
    filecolor=black,
    urlcolor=black
}

\title{Semantic Causal-Factor Inference from Aviation Incident Narratives Using a Variational Autoencoder with Cosine-Similarity-Based Reconstruction}

\author{
Aziida Nanyonga$^{1,*}$,
Hassan Wasswa$^{2}$,
Ugur Turhan$^{3}$,
Keith Francis Joiner$^{4}$,
and Graham Wild$^{3}$
}

\date{}

\begin{document}

\begin{abstract}
Aviation accident and incident investigations generate extensive unstructured textual information containing evidence relevant to the causes and contributing factors of safety occurrences. Automatically extracting such information is challenging because causal evidence may be distributed across long and complex investigation narratives. This study proposes a semantic causal-factor inference framework combining natural language processing with a variational autoencoder (VAE) to learn the relationship between aviation investigation narratives and expert-reported probable causes. Investigation narratives and their corresponding probable causes are transformed into numerical representations, after which the encoder maps narrative representations to a probabilistic latent space. The decoder estimates representations of the corresponding probable causes and is trained using an objective that combines Kullback--Leibler divergence with cosine-similarity-based semantic reconstruction. The framework was evaluated using 20,919 finalized U.S. National Transportation Safety Board investigation reports from 2005--2020. On the held-out test set, the predicted and expert-reported probable-cause representations achieved a mean cosine similarity of 0.786 (SD = 0.120). The predicted representations also yielded interpretable terms associated with causal information in the reports. The results demonstrate the potential of probabilistic latent representation learning for AI-assisted extraction of causal information from aviation safety narratives while retaining expert investigation as the basis for formal causal determination.
\end{abstract}

\maketitle

\begin{center}
\small
$^{1}$SDAIA-KFUPM Joint Research Center for Artificial Intelligence,
King Fahd University of Petroleum and Minerals (KFUPM),
Dhahran 31261, Saudi Arabia

\medskip
$^{2}$School of Systems and Computing,
University of New South Wales,
Canberra, ACT 2600, Australia

\medskip
$^{3}$School of Science,
University of New South Wales,
Canberra, ACT 2612, Australia

\medskip
$^{4}$Capability Systems Centre,
University of New South Wales,
Canberra, ACT 2600, Australia

\medskip
$^{*}$Corresponding author:
\href{mailto:aziida.nanyonga@kfupm.edu.sa}{aziida.nanyonga@kfupm.edu.sa}
\end{center}

\medskip


\bigskip

\section{Introduction}
\label{sec:introduction}
Aviation accident and incident investigation is fundamental to identifying safety deficiencies and preventing the recurrence of adverse events. Investigation reports integrate evidence concerning aircraft systems, operational conditions, human actions, environmental factors, and the sequence of events surrounding an occurrence. Flight recorders provide particularly important evidence during this process \cite{vidovic2022importance}. However, substantial investigative information and causal reasoning are ultimately documented in textual form. As aviation safety databases continue to grow, the volume and complexity of these narratives make systematic large-scale manual analysis increasingly challenging.

Natural language processing (NLP) provides computational methods for transforming unstructured text into representations that can be analyzed using machine-learning algorithms \cite{bird2009natural,jurafsky2024speech}. Within aviation safety research, NLP and text-mining techniques have been applied to identify latent themes, classify safety occurrences, and extract information from incident and accident narratives \cite{paul2018nlp}. Structural topic modeling has been used to identify latent topics and temporal trends in aviation incident reports \cite{kenneth2018}, while topic-modeling and clustering approaches have been investigated for discovering patterns in aviation occurrence narratives \cite{nanyonga2025socrata}. Deep-learning methods have also been applied to aviation safety classification problems, including causal-factor identification \cite{dong2021identifying, nanyonga2025classification}, sequential classification of safety occurrences and investigation of the effects of dataset size and data source on aviation-safety prediction models \cite{nanyonga2023impact}. Collectively, these studies demonstrate that useful safety information can be learned from unstructured aviation narratives.

Inferring causal information from an investigation narrative, however, differs from assigning a report to a predefined category or discovering its dominant topic. An investigator-reported probable cause may depend on information distributed throughout a long narrative and may reflect interactions among technical failures, human actions, operational conditions, and environmental factors. Consequently, causal information is not necessarily represented by an individual keyword, a single sentence, or a predefined class. This motivates an alternative formulation in which the relationship between an investigation narrative and its corresponding probable cause is learned directly within a continuous semantic representation space.

Distributed word representations provide one mechanism for constructing such a semantic space. Global Vectors for Word Representation (GloVe), for example, learns vector representations from word co-occurrence statistics such that semantic relationships can be represented geometrically \cite{pennington2014glove}. More recent contextual architectures, including BERT, learn representations that depend on the surrounding linguistic context \cite{devlin2018bert}. Although contextual language models provide capabilities beyond static word embeddings, the present study investigates whether causal information can be learned using latent-variable architecture based on GloVe representations and a variational autoencoder (VAE).

VAEs extend conventional autoencoders by representing observations through probability distributions in a continuous latent space \cite{Kingma2013AutoEncodingVB}. This probabilistic formulation enables structured representation learning while retaining a generative decoding mechanism. VAEs have been investigated for natural-language applications including sentence representation and generation \cite{bowman2015generating,song2019latent}, and semantic-similarity information has previously been incorporated into VAE-based text representation models \cite{su2018integrating}. VAE-based architectures have also been investigated in aviation-safety applications, including unsupervised anomaly detection in high-dimensional flight data using convolutional variational autoencoders \cite{memarzadeh2020unsupervised}. This demonstrates the applicability of probabilistic latent-variable models to aviation data, although such work focuses on operational time-series anomaly detection rather than semantic inference from investigation narratives. 

In this study, we investigate a VAE-based framework for learning this narrative--cause relationship. Rather than treating causal-factor inference exclusively as a classification problem over a predefined set of labels, the proposed framework learns from paired investigation narratives and probable-cause statements. The investigation narrative is encoded into a probabilistic latent representation, from which the decoder estimates a representation associated with the corresponding probable cause. To encourage semantic agreement between the predicted and expert-reported causal representations, cosine similarity is incorporated into the reconstruction component of the training objective.

The framework is evaluated using 20,919 finalized aviation accident and incident investigation records obtained from the U.S. National Transportation Safety Board (NTSB) for the period 2005--2020. Each record contains an investigation analysis narrative and the corresponding probable-cause statement established during the investigation. This paired structure enables the model to learn a continuous semantic relationship between descriptive occurrence information and expert-reported causal information.

The principal contributions of this study are as follows:
\begin{itemize}
    \item Aviation causal-factor inference is formulated as a semantic representation-learning problem in which investigation narratives are mapped to representations of expert-reported probable causes.
    
    \item A VAE-based architecture is developed to learn a probabilistic latent representation connecting aviation investigation narratives with their corresponding causal information.
    
    \item Cosine-similarity-based semantic reconstruction is incorporated into the VAE training objective to encourage semantic agreement between predicted and expert-reported probable-cause representations.
    
    \item The proposed framework is evaluated using 20,919 finalized NTSB investigation reports, with performance assessed through the semantic similarity between predicted and expert-reported causal representations.
    
    \item The predicted representations are examined through associated causal terms to provide an interpretable indication of the causal information learned by the model.
\end{itemize}

The remainder of this paper is organized as follows. Section~\ref{sec:relatedwork} reviews previous research on NLP for aviation safety analysis, VAE-based representation learning, and AI-based causal analysis. Section~\ref{sec:methods} describes the dataset, text preprocessing and numerical representation, VAE architecture, semantic-similarity-based training objective, inference procedure, and experimental design. Section~\ref{sec:results} presents the experimental results. Section~\ref{sec:discussion} discusses the findings, their implications, and the limitations of the proposed approach. Finally, Section~\ref{sec:conclusions} summarizes the main conclusions and directions for future research.

\section{Related Work}
\label{sec:relatedwork}
\subsection{Natural Language Processing for Aviation Safety Analysis}

Aviation safety databases contain substantial amounts of unstructured textual information, including accident and incident narratives, investigator analyses, and safety reports. Consequently, natural language processing (NLP) has increasingly been used to extract patterns and safety-relevant information that would otherwise require extensive manual review. Applications of NLP to aviation safety have commonly formulated report analysis as topic discovery, classification, or information extraction.

Topic modeling provides an unsupervised approach for identifying latent thematic structure within collections of textual documents \cite{jelodar2019latent}. In aviation safety research, Kuhn applied structural topic modeling to aviation incident reports to identify latent topics and temporal trends \cite{kenneth2018}. Topic-modeling approaches have subsequently been investigated for discovering patterns within aviation occurrence narratives. Nanyonga and Wild compared latent Dirichlet allocation (LDA), non-negative matrix factorization (NMF), and probabilistic latent semantic analysis (PLSA) for analyzing aviation safety narratives \cite{nanyonga2025socrata}. More recent work has extended this direction through semantic topic modeling, including comparative analysis of BERTopic and PLSA for extracting themes from aviation safety reports \cite{nanyonga2025bertopic}.

Deep-learning approaches have also demonstrated that distributed textual representations can support supervised aviation safety analysis. Dong et al. proposed a deep-learning approach for identifying incident causal factors \cite{dong2021identifying}, while recurrent architectures have been applied to phase-of-flight and sequential occurrence classification \cite{goblet2015identifying}. Comparative studies have further investigated deep-learning architectures for textual aviation-safety classification \cite{nanyonga2025deep}, while the influence of dataset size and data source on aviation-safety prediction has also been examined \cite{nanyonga2023impact}. NLP and deep-learning models have additionally been applied to predicting injury levels from Australian aviation incident reports \cite{nanyonga2025injury} and to post-accident analysis classification \cite{nanyonga2025postaccident}. More recently, Transformer-based deep-learning models have been investigated for flight-phase classification from Australian Transport Safety Bureau aviation safety reports \cite{nanyonga2026flightphase}. Collectively, these studies demonstrate the growing use of both conventional NLP methods and neural language representations for extracting structured safety information from aviation narratives.

\subsection{Variational Autoencoders and Semantic Representation Learning}

Variational autoencoders (VAEs) provide a probabilistic latent-variable framework in which high-dimensional observations are represented within a continuous latent space \cite{Kingma2013AutoEncodingVB}. Unlike deterministic autoencoders, a VAE learns the parameters of a latent probability distribution and regularizes that distribution toward a specified prior. This formulation supports continuous representation learning while retaining a decoding mechanism through which information can be reconstructed from latent samples.

VAEs have been investigated extensively for natural-language representation and generation. Bowman et al. demonstrated that sentences can be represented within a continuous latent space using a VAE architecture \cite{bowman2015generating}. Subsequent work has investigated expanded latent spaces for sentence generation \cite{song2019latent} and the integration of topic information into VAE architectures for text semantic similarity \cite{su2018integrating}. These studies indicate that probabilistic latent-variable models can encode semantic properties of textual data rather than functioning solely as dimensionality-reduction mechanisms.

Within aviation applications, variational autoencoders have been investigated primarily for representation learning and anomaly detection in operational data rather than textual investigation narratives. Memarzadeh et al. developed a convolutional variational autoencoder for unsupervised anomaly detection in high-dimensional aviation time-series data and demonstrated the approach using commercial-flight take-off data \cite{memarzadeh2020unsupervised}. VAE-based approaches have also been investigated for anomaly detection in aviation-related operational data \cite{luo2021adsb}. These studies demonstrate the applicability of VAE-based latent representation learning to aviation-safety data; however, their tasks differ fundamentally from the present study because they concern anomaly detection in operational flight parameters rather than semantic mapping between investigation narratives and investigator-reported causal information.

An additional consideration in textual representation learning is the reconstruction objective. Conventional reconstruction losses quantify numerical differences between target and reconstructed representations. When the target represents textual information, however, numerical reconstruction error alone may not adequately describe semantic agreement. Cosine similarity provides a geometric measure of the alignment between distributed representations, and previous work has investigated semantic-similarity information within VAE-based textual models \cite{su2018integrating}. This provides the methodological motivation for incorporating semantic agreement into the reconstruction objective used in the present study.

\subsection{Recent AI Approaches to Aviation Causal Analysis}

Recent research has moved beyond general report classification toward explicit causal-information extraction and reasoning from aviation narratives. Ren et al. proposed a retrieval-augmented-generation-aided causal identification framework that combines retrieval from historical aviation accident reports with large language model (LLM) processing \cite{ren2025rag}. This formulation uses retrieved domain information to support causal identification rather than learning a dedicated probabilistic mapping between paired narrative and probable-cause representations.

Liu et al. investigated LLM-based accident reasoning using the Human Factors Analysis and Classification System (HFACS) to guide chain-of-thought reasoning \cite{liu2025hfacs}. Their framework focuses specifically on inferring unsafe acts and associated preconditions from general-aviation witness narratives. This work demonstrates the potential benefit of integrating established aviation-safety knowledge with foundation-model reasoning.

More recently, Chen et al. proposed the Aviation Reports Causality Extraction Model (ARCEM), which formulates causality extraction from NTSB causal narratives as a sequence-labeling problem \cite{chen2026arcem}. The architecture combines contextual BERT representations, multi-granular textual features, a Transformer, and a conditional random field decoder to identify cause, effect, and embedded-causality spans. This formulation provides explicit causal structures from text but differs from the present task of estimating the semantic representation of an investigator-reported probable cause from a longer investigation narrative.

Lin et al. proposed AI4AirSafe, a hybrid knowledge-graph--LLM framework for aviation accident causal reasoning \cite{lin2026ai4airsafe}. Their approach represents events and temporal information within an aviation knowledge graph and combines structured retrieval with LLM reasoning to reconstruct causal chains and generate evidence-grounded explanations. Such approaches illustrate the increasing emphasis on explicit structure, contextual language models, and explainability in AI-assisted aviation accident analysis.

\subsection{Research Gap and Positioning of the Present Study}

The existing literature reflects several complementary formulations of aviation safety text analysis. Topic-modeling approaches identify latent themes within collections of reports \cite{kenneth2018,nanyonga2025bertopic}; supervised deep-learning methods assign reports to predefined categories or safety outcomes \cite{dong2021identifying,nanyonga2025deep,nanyonga2025injury,nanyonga2025postaccident,nanyonga2026flightphase}; recent Transformer architectures extract explicit causal spans \cite{chen2026arcem}; and LLM-based approaches perform causal identification or reasoning using retrieved information, domain frameworks, or structured knowledge \cite{ren2025rag,liu2025hfacs,lin2026ai4airsafe}.

The present study investigates a different formulation. Given paired observations consisting of an investigation narrative and an expert-reported probable cause, the objective is to learn a continuous probabilistic representation connecting the two. The causal output is therefore neither restricted to a predefined classification label nor produced through unrestricted LLM text generation. Instead, an investigation narrative is encoded into a latent probability distribution from which the model estimates a semantic representation associated with the corresponding probable cause.

The proposed approach consequently examines whether a compact VAE-based latent-variable model can capture the semantic relationship between long-form aviation investigation narratives and investigator-established causal information. Cosine-similarity-based reconstruction is incorporated into the training objective to encourage alignment between the predicted and target causal representations. The resulting representation can subsequently be examined through associated causal terms, providing an interpretable indication of the causal information captured by the model.

\section{Materials and Methods}
\label{sec:methods}

\subsection{Study Framework and Problem Formulation}
\label{sec:framework}

The proposed framework learns a mapping between an aviation investigation narrative and the corresponding expert-reported probable cause. Let the dataset be represented as
\begin{equation}
\mathcal{D}=\{(S_i,C_i)\}_{i=1}^{N},
\label{eq:dataset}
\end{equation}
where $S_i$ denotes the analysis narrative for the $i$th aviation occurrence, $C_i$ denotes its corresponding probable-cause statement, and $N$ is the total number of investigation records.

After text preprocessing and numerical representation, the investigation narrative is represented by $\mathbf{x}_i$, while the corresponding probable cause is represented by $\mathbf{y}_i$. The objective is to learn a mapping from $\mathbf{x}_i$ to an estimated causal representation $\hat{\mathbf{y}}_i$. Rather than assigning the occurrence to a predefined causal category, the proposed method learns this relationship within a continuous latent representation space.

The model consists of an encoder and a decoder. The encoder maps the narrative representation to a probabilistic latent representation $\mathbf{z}_i$, and the decoder maps the latent representation toward an estimate of the corresponding probable-cause representation. The overall mapping is expressed as
\begin{equation}
\mathbf{x}_i
\xrightarrow{q_{\phi}(\mathbf{z}\mid\mathbf{x}_i)}
\mathbf{z}_i
\xrightarrow{p_{\theta}(\mathbf{y}\mid\mathbf{z}_i)}
\hat{\mathbf{y}}_i,
\label{eq:overall_mapping}
\end{equation}
where $q_{\phi}(\mathbf{z}\mid\mathbf{x}_i)$ denotes the encoder parameterized by $\phi$, $p_{\theta}(\mathbf{y}\mid\mathbf{z}_i)$ denotes the decoder parameterized by $\theta$, and $\hat{\mathbf{y}}_i$ denotes the predicted representation associated with the probable cause.

Training combines regularization of the probabilistic latent space with semantic agreement between the predicted and expert-reported probable-cause representations. The complete workflow therefore consists of: (1) acquisition of finalized aviation investigation records; (2) preprocessing and numerical representation of the investigation narratives and probable causes; (3) encoding of each narrative representation into a latent probability distribution; (4) sampling of a latent representation and decoding it toward the corresponding probable-cause representation; and (5) optimization and evaluation using semantic similarity between the predicted and target causal representations.

\begin{figure}[H]
\centering
\includegraphics[width=\textwidth]{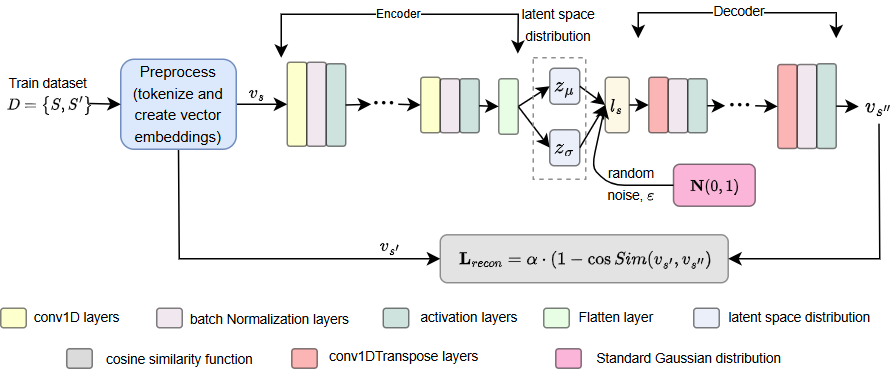}
\caption{Proposed framework for semantic causal-factor inference from aviation investigation narratives. The investigation narrative is preprocessed and numerically represented before being encoded into a latent representation. The decoder estimates the representation of the corresponding probable cause, and semantic similarity between the predicted and expert-reported representations is incorporated into model training.}
\label{research_flow}
\end{figure}

Figure~\ref{research_flow} presents the overall architecture and information flow of the proposed approach.

\subsection{Dataset}
\label{sec:dataset}

The dataset used in this study was obtained from finalized aviation accident and incident investigation records maintained by the U.S. National Transportation Safety Board (NTSB). Investigation records covering the period from 2005 to 2020 were considered. After extraction of the textual fields required for the proposed task, the dataset comprised 20,919 investigation records.

For each occurrence, two textual fields were used: the \textit{analysis narrative} and the corresponding \textit{probable cause}. The analysis narrative contains the investigators' description and analysis of the occurrence, whereas the probable-cause field records the causal determination reported following the investigation. The analysis narrative was used as the model input, and the corresponding probable-cause statement was used as the target.

The analysis narratives were longer on average than the probable-cause statements. The narratives contained an average of 1116 words (SD = 858.36), whereas the probable-cause statements contained an average of 165 words (SD = 93.12). Maximum sequence lengths of 1200 and 200 were used for the analysis narratives and probable-cause statements, respectively.

Table~\ref{tab:dataset} summarizes the principal characteristics of the dataset.

\begin{table}[H]
\caption{Characteristics of the NTSB aviation investigation dataset used in this study.}
\label{tab:dataset}
\begin{tabularx}{\textwidth}{LC}
\toprule
\textbf{Characteristic} & \textbf{Value} \\
\midrule
Data source & U.S. National Transportation Safety Board (NTSB) \\
Investigation status & Finalized reports \\
Study period & 2005--2020 \\
Number of records & 20,919 \\
Model input & Analysis narrative \\
Model target & Probable cause \\
Mean narrative length & 1116 words \\
Narrative-length SD & 858.36 words \\
Mean probable-cause length & 165 words \\
Probable-cause-length SD & 93.12 words \\
Maximum narrative sequence length & 1200 \\
Maximum probable-cause sequence length & 200 \\
\bottomrule
\end{tabularx}
\end{table}

\subsection{Text Preprocessing and Numerical Representation}
\label{sec:preprocessing}

The analysis narratives and probable-cause statements were preprocessed before numerical representation and model training. The preprocessing pipeline was implemented using spaCy and the Natural Language Toolkit (NLTK) \cite{bird2009natural}. The procedure included tokenization and removal of stop words, punctuation, URLs, and HTML-related content. Stemming and lemmatization were also applied to reduce lexical variation in the textual data.

Following preprocessing, the textual data were transformed into numerical representations using pretrained Global Vectors for Word Representation (GloVe) embeddings \cite{pennington2014glove}. GloVe represents words within a continuous vector space learned from global word co-occurrence statistics, allowing words with related semantic characteristics to occupy nearby regions of the representation space.

The resulting numerical representations were subsequently normalized using min--max scaling. For a numerical value $x_i$, the normalized value $x_i'$ was calculated as

\begin{equation}
x_i' =
\frac{x_i-x_{\min}}
{x_{\max}-x_{\min}},
\label{eq:minmax}
\end{equation}

where $x_{\min}$ and $x_{\max}$ denote the minimum and maximum values, respectively, of the numerical representation being normalized.

To obtain fixed-length inputs for the neural-network architecture, the analysis-narrative representations were restricted to a maximum sequence length of 1200, while the probable-cause representations were restricted to a maximum sequence length of 200. The resulting narrative and probable-cause representations were provided to the model with dimensions $(1200,1)$ and $(200,1)$, respectively. The narrative representation served as the encoder input, whereas the probable-cause representation served as the target for decoder reconstruction.

\subsection{Variational Autoencoder Formulation}
\label{sec:vae}

A variational autoencoder learns a probabilistic latent representation of the input rather than a deterministic encoding \cite{Kingma2013AutoEncodingVB}. Given an input narrative representation $\mathbf{x}$, the encoder approximates the posterior distribution of the latent variable $\mathbf{z}$ as

\begin{equation}
q_{\phi}(\mathbf{z}\mid\mathbf{x})
=
\mathcal{N}
\left(
\boldsymbol{\mu},
\operatorname{diag}(\boldsymbol{\sigma}^{2})
\right),
\label{eq:posterior}
\end{equation}

where $\boldsymbol{\mu}$ and $\boldsymbol{\sigma}$ represent the mean and standard deviation of the latent distribution estimated by the encoder.

A latent representation is sampled using the reparameterization procedure

\begin{equation}
\mathbf{z}
=
\boldsymbol{\mu}
+
\boldsymbol{\sigma}\odot\boldsymbol{\epsilon},
\qquad
\boldsymbol{\epsilon}\sim\mathcal{N}(\mathbf{0},\mathbf{I}),
\label{eq:reparameterization}
\end{equation}

where $\odot$ denotes element-wise multiplication. This formulation enables gradient-based optimization through the stochastic sampling operation.

The latent distribution is regularized toward a standard multivariate Gaussian prior,

\begin{equation}
p(\mathbf{z})=\mathcal{N}(\mathbf{0},\mathbf{I}),
\label{eq:prior}
\end{equation}

using the Kullback--Leibler (KL) divergence between the approximate posterior and the prior. For a Gaussian latent distribution, the KL-divergence term can be expressed as

\begin{equation}
\mathcal{L}_{\mathrm{KL}}
=
-\frac{1}{2}
\sum_{j=1}^{d}
\left(
1+\log \sigma_j^{2}
-\mu_j^{2}
-\sigma_j^{2}
\right),
\label{eq:kl}
\end{equation}

where $d$ denotes the dimensionality of the latent space.

The sampled latent representation $\mathbf{z}$ is subsequently provided to the decoder, which estimates the representation of the corresponding probable cause,

\begin{equation}
\hat{\mathbf{y}}
=
f_{\theta}(\mathbf{z}),
\label{eq:decoder}
\end{equation}

where $f_{\theta}$ denotes the decoder parameterized by $\theta$ and $\hat{\mathbf{y}}$ denotes the predicted probable-cause representation.

\subsection{Model Architecture}
\label{sec:architecture}

Following hyperparameter selection, different encoder--decoder configurations were evaluated by varying the number of convolutional filters, kernel sizes, and strides. The final architecture consists of a one-dimensional convolutional encoder and a transposed-convolutional decoder. Table~\ref{vae_model_structure} summarizes the architecture used for the final experiments.

The encoder receives an analysis-narrative representation with shape $(1200,1)$. A sequence of one-dimensional convolutional layers progressively transforms the narrative representation before the resulting features are flattened and projected to a 20-dimensional latent representation. The decoder receives the 20-dimensional latent representation and first projects it through a fully connected layer containing 64 units. After reshaping, a sequence of one-dimensional transposed-convolutional layers progressively reconstructs the target representation until an output shape of $(200,1)$ is obtained, corresponding to the representation of the probable-cause sequence.

LeakyReLU activation was used in the convolutional hidden layers of the encoder--decoder architecture. The final architecture therefore performs an asymmetric narrative-to-cause mapping: the encoder operates on the longer $(1200,1)$ investigation-narrative representation, whereas the decoder produces the shorter $(200,1)$ probable-cause representation.

\begin{table}[H]
\caption{Final VAE encoder and decoder architecture used in the experiments, including the number of trainable parameters in each layer.}
\label{vae_model_structure}
\centering
\resizebox{\textwidth}{!}{%
\begin{tabular}{lccccr|lccccr}
\toprule
\multicolumn{6}{c|}{\textbf{Encoder}} &
\multicolumn{6}{c}{\textbf{Decoder}} \\
\midrule
\textbf{Layer} &
\textbf{Filters/Units} &
\textbf{Output Shape} &
\textbf{Activation} &
\textbf{Kernel Size} &
\textbf{Parameters} &
\textbf{Layer} &
\textbf{Filters/Units} &
\textbf{Output Shape} &
\textbf{Activation} &
\textbf{Kernel Size} &
\textbf{Parameters} \\
\midrule

Input
& -- & $(1200,1)$ & -- & -- & 0
&
Input
& 20 & $(20,)$ & -- & -- & 0 \\

--
& -- & -- & -- & -- & 0
&
Dense
& 64 & $(64,)$ & LeakyReLU & -- & 1,344 \\

--
& -- & -- & -- & -- & 0
&
Reshape
& -- & $(1,64)$ & -- & -- & 0 \\

Conv1D
& 256 & $(134,256)$ & LeakyReLU & 36 & 9,472
&
Conv1DTranspose
& 64 & $(3,64)$ & LeakyReLU & 4 & 16,448 \\

Conv1D
& 128 & $(17,128)$ & LeakyReLU & 25 & 819,328
&
Conv1DTranspose
& 64 & $(6,64)$ & LeakyReLU & 4 & 16,448 \\

Conv1D
& 128 & $(4,128)$ & LeakyReLU & 25 & 409,728
&
Conv1DTranspose
& 64 & $(19,64)$ & LeakyReLU & 4 & 16,448 \\

Conv1D
& 64 & $(1,64)$ & LeakyReLU & 16 & 131,136
&
Conv1DTranspose
& 64 & $(22,64)$ & LeakyReLU & 4 & 16,448 \\

Conv1D
& 64 & $(1,64)$ & LeakyReLU & 9 & 36,928
&
Conv1DTranspose
& 64 & $(47,64)$ & LeakyReLU & 5 & 20,544 \\

Conv1D
& 64 & $(1,64)$ & LeakyReLU & 4 & 16,448
&
Conv1DTranspose
& 64 & $(47,64)$ & LeakyReLU & 4 & 16,448 \\

Conv1D
& 64 & $(1,64)$ & LeakyReLU & 2 & 8,256
&
Conv1DTranspose
& 64 & $(96,64)$ & LeakyReLU & 4 & 16,448 \\

Conv1D
& 64 & $(1,64)$ & LeakyReLU & 4 & 16,448
&
Conv1DTranspose
& 64 & $(194,64)$ & LeakyReLU & 4 & 16,448 \\

Conv1D
& 64 & $(1,64)$ & LeakyReLU & 4 & 16,448
&
Conv1DTranspose
& 64 & $(197,64)$ & LeakyReLU & 3 & 12,352 \\

Conv1D
& 64 & $(1,64)$ & LeakyReLU & 2 & 8,256
&
Conv1DTranspose
& 64 & $(199,64)$ & LeakyReLU & 2 & 8,256 \\

Conv1D
& 64 & $(1,64)$ & LeakyReLU & 1 & 4,160
&
Conv1DTranspose
& 64 & $(200,64)$ & LeakyReLU & 1 & 4,160 \\

Flatten
& -- & -- & -- & -- & 0
&
Conv1DTranspose
& 1 & $(200,1)$ & LeakyReLU & 1 & 65 \\

Dense
& 20 & -- & LeakyReLU & -- & 1,300
&
Output
& -- & $(200,1)$ & -- & -- & 0 \\

\midrule
\multicolumn{5}{r}{\textbf{Encoder total:}} &
\textbf{1,477,908}
&
\multicolumn{5}{r}{\textbf{Decoder total:}} &
\textbf{161,857} \\

\midrule
\multicolumn{10}{r}{\textbf{Total trainable parameters:}} &
\multicolumn{1}{r}{\textbf{1,639,765}} \\

\bottomrule
\end{tabular}%
}
\end{table}

\subsection{Semantic-Similarity-Based Training Objective}
\label{sec:loss}

The training objective combines regularization of the latent distribution with a semantic reconstruction component. Latent-space regularization is achieved using the Kullback--Leibler (KL) divergence between the approximate posterior distribution learned by the encoder and the standard Gaussian prior, as defined in Equation~\eqref{eq:kl}.

Because the decoder is optimized to estimate a representation of the expert-reported probable cause, semantic agreement between the predicted and target representations is measured using cosine similarity. Given the target probable-cause representation $\mathbf{y}$ and the predicted representation $\hat{\mathbf{y}}$, cosine similarity is defined as

\begin{equation}
\mathrm{cosSim}
(\mathbf{y},\hat{\mathbf{y}})
=
\frac{
\mathbf{y}^{\mathrm{T}}\hat{\mathbf{y}}
}{
\|\mathbf{y}\|_{2}
\|\hat{\mathbf{y}}\|_{2}
},
\label{eq:cosine}
\end{equation}

where $\|\cdot\|_{2}$ denotes the Euclidean norm.

The semantic reconstruction loss is defined as

\begin{equation}
\mathcal{L}_{\mathrm{sem}}
=
\alpha
\left[
1-
\mathrm{cosSim}
(\mathbf{y},\hat{\mathbf{y}})
\right],
\label{eq:semantic_loss}
\end{equation}

where $\alpha$ is a scaling coefficient controlling the contribution of the semantic reconstruction component.

The overall training objective combines the KL-divergence and semantic reconstruction terms:

\begin{equation}
\mathcal{L}_{\mathrm{total}}
=
\mathcal{L}_{\mathrm{KL}}
+
\mathcal{L}_{\mathrm{sem}}.
\label{eq:total_loss}
\end{equation}

Minimizing $\mathcal{L}_{\mathrm{KL}}$ regularizes the learned latent distribution toward the assumed Gaussian prior, whereas minimizing $\mathcal{L}_{\mathrm{sem}}$ increases the cosine similarity between the predicted and expert-reported probable-cause representations. The combined objective therefore encourages both a regularized latent representation and semantic agreement between the decoder output and the target causal representation.

\subsection{Causal-Factor Inference}
\label{sec:inference}

After training, the model can be applied to an investigation narrative without providing its corresponding probable-cause statement. The narrative is first processed using the same text-preprocessing and numerical-representation procedure employed during model training. Let $s_i$ denote the textual narrative associated with the $i$th occurrence. The narrative is transformed into its numerical representation $\mathbf{v}_i$, which is subsequently supplied to the trained encoder.

The encoder maps $\mathbf{v}_i$ to the parameters of the learned latent distribution, from which the latent representation $\mathbf{z}_i$ is obtained. The decoder then uses this representation to estimate a numerical representation $\hat{\mathbf{v}}_i$ associated with the probable cause.

\begin{equation}
s_i
\rightarrow
\mathbf{v}_i
\rightarrow
q_{\phi}(\mathbf{z}\mid\mathbf{v}_i)
\rightarrow
\mathbf{z}_i
\rightarrow
p_{\theta}(\mathbf{y}\mid\mathbf{z}_i)
\rightarrow
\hat{\mathbf{v}}_i.
\label{eq:inference}
\end{equation}

Because the decoder output is a numerical representation rather than directly interpretable natural-language text, the predicted representation $\hat{\mathbf{v}}_i$ is mapped back to the textual representation space using the inverse of the numerical text-representation procedure applied during preprocessing. This produces a sequence of words, denoted by $\hat{s}_i$, that can be examined by a human analyst:

\begin{equation}
\hat{\mathbf{v}}_i
\rightarrow
\hat{s}_i.
\label{eq:inverse_mapping}
\end{equation}

The complete inference sequence is therefore

\begin{equation}
s_i
\rightarrow
\mathbf{v}_i
\rightarrow
\mathbf{z}_i
\rightarrow
\hat{\mathbf{v}}_i
\rightarrow
\hat{s}_i.
\label{eq:complete_inference}
\end{equation}

The resulting textual output should be interpreted as a collection of indicative causal terms rather than as a complete investigator-style probable-cause statement. The framework estimates causal information learned from historical narrative--probable-cause relationships and is intended to support expert analysis rather than independently establish the official cause of an aviation occurrence.

\begin{figure}[H]
    \centering
     \includegraphics[width=0.95\textwidth]{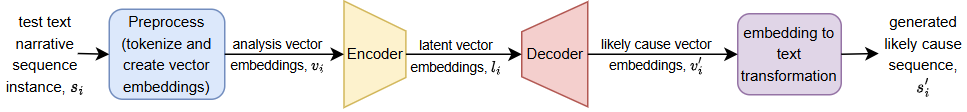}
    \caption{Inference procedure for estimating causal information from an aviation investigation narrative. The narrative sequence $s_i$ is converted into a numerical representation $\mathbf{v}_i$ and processed by the encoder to obtain the latent representation $\mathbf{z}_i$. The decoder produces the predicted causal representation $\hat{\mathbf{v}}_i$, which is subsequently mapped back to the textual representation space to obtain interpretable causal terms $\hat{s}_i$.}
    \label{fig:inference}
\end{figure}

Figure~\ref{fig:inference} illustrates the inference process.

\subsection{Experimental Design}
\label{sec:experimental}

The dataset was randomly divided into model-development and test subsets. A total of 85\% of the observations were allocated to model development, while the remaining 15\% were retained as a held-out test set. Within the model-development subset, 10\% of the observations were reserved for validation. The validation subset was used during model development and hyperparameter selection, whereas the held-out test subset was reserved for evaluation of the selected model configuration.

Model development was performed in Python using TensorFlow, together with supporting libraries including NLTK, scikit-learn, NumPy, and pandas. The experiments were conducted on an Ubuntu-based computing server equipped with 256 CPU cores, four NVIDIA A100-SXM4 GPUs with 80~GB of memory per GPU, and 512~GB of system memory.

Hyperparameter exploration included the optimization algorithm, learning rate, batch size, vocabulary size, and latent-space dimensionality. Adam, Nadam, and RMSprop were evaluated as optimization algorithms, with learning rates of 0.01, 0.001, and 0.0001. Adam \cite{kingma2014adam} and Nadam \cite{dozat2016incorporating} were included among the adaptive optimization approaches evaluated during model development.

Additional experiments considered vocabulary sizes ranging from 10,000 to 150,000, batch sizes ranging from 8 to 64, \( \alpha= \)\{500, 600, \( \ldots\), 1500\}, and latent-space dimensions ranging from 10 to 100. Based on the model-development experiments, the final configuration used the Adam optimizer with a learning rate of 0.001, \(\alpha = 1000\), a batch size of 16, and a latent-space dimensionality of 20.

The selected model was subsequently evaluated on the held-out test subset using cosine similarity between the predicted causal representation and the corresponding expert-reported probable-cause representation.

\section{Results}
\label{sec:results}

\subsection{Hyperparameter and Architecture Selection}
\label{sec:hyperparameter_results}

The effects of the optimization algorithm, learning rate, vocabulary size, batch size, latent-space dimensionality, and encoder--decoder configuration were investigated during model development. Initial experiments evaluated Adam, Nadam, and RMSprop optimizers at different learning rates. Performance was measured using cosine similarity between the predicted and expert-reported probable-cause representations.

Adam achieved its highest mean cosine similarity at a learning rate of 0.001 ($M=0.663$, SD $=0.150$), whereas Nadam achieved the same mean similarity at a learning rate of 0.01 ($M=0.663$, SD $=0.156$). RMSprop achieved its highest mean similarity of 0.649 at a learning rate of 0.01. Adam with a learning rate of 0.001 was selected for the subsequent hyperparameter experiments. Table~\ref{tab:optimizer_lr} summarizes the optimizer and learning-rate results.

\begin{table}[H]
\caption{Optimizer and learning-rate evaluation during model development.}
\label{tab:optimizer_lr}
\centering
\begin{tabular}{lccc}
\toprule
\textbf{Optimizer} &
\textbf{Learning Rate} &
\textbf{Mean Cosine Similarity} &
\textbf{SD} \\
\midrule
Adam
& 0.0100 & 0.623 & 0.126 \\
& 0.0010 & \textbf{0.663} & 0.150 \\
& 0.0001 & 0.644 & 0.163 \\
\midrule
Nadam
& 0.0100 & \textbf{0.663} & 0.156 \\
& 0.0010 & 0.661 & 0.140 \\
& 0.0001 & 0.644 & 0.158 \\
\midrule
RMSprop
& 0.0100 & 0.649 & 0.131 \\
& 0.0010 & 0.634 & 0.181 \\
& 0.0001 & 0.641 & 0.162 \\
\bottomrule
\end{tabular}
\end{table}

Following optimizer and learning-rate exploration, additional experiments varied the maximum vocabulary size from 10,000 to 150,000, batch size from 8 to 64, and latent-space dimensionality from 10 to 100. Across the evaluated hyperparameter combinations, average cosine-similarity values ranged from approximately 0.626 to 0.712.

The configuration consisting of the Adam optimizer, a learning rate of 0.001, a batch size of 16, and a latent-space dimensionality of 20 produced the highest average similarity among the evaluated hyperparameter combinations and was selected for subsequent model development.

The encoder--decoder architecture was then further investigated by varying convolutional parameters, including the number of filters, strides, and kernel sizes. The final architecture was selected from these experiments and used for the test-set evaluation.

\subsection{Overall Test-Set Performance}
\label{sec:test_results}

The selected model was evaluated using the held-out test subset. For each test observation, cosine similarity was calculated between the model-predicted causal representation and the corresponding expert-reported probable-cause representation.

Across the test observations, the model achieved a mean cosine similarity of

\begin{equation}
\overline{S}_{\mathrm{cos}} = 0.786,
\label{eq:mean_cosine}
\end{equation}

with a standard deviation of

\begin{equation}
\sigma_{\mathrm{cos}} = 0.120.
\label{eq:sd_cosine}
\end{equation}

\begin{figure}[H]
    \centering
    \includegraphics[width=0.65\textwidth]{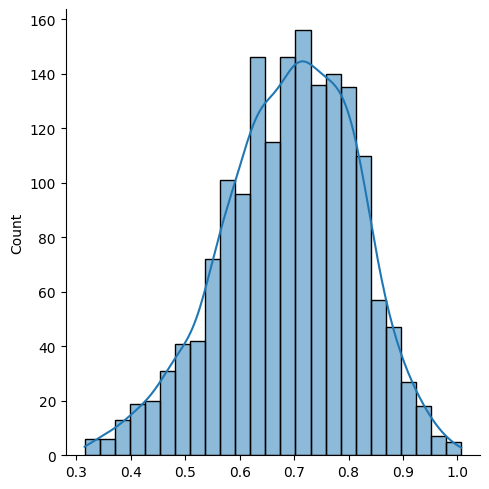}
    \caption{Distribution of cosine-similarity scores between the predicted causal representations and the corresponding expert-reported probable-cause representations for the held-out test observations.}
    \label{normal_plot}
\end{figure}

Figure~\ref{normal_plot} shows the distribution of cosine-similarity scores obtained for the test observations. The distribution is concentrated toward higher similarity values, with relatively few observations producing scores below 0.5.

\begin{figure}[H]
    \centering
    \includegraphics[width=0.75\textwidth]{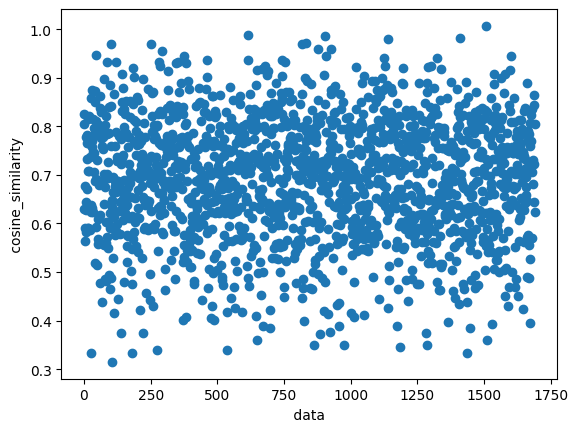}
    \caption{Cosine-similarity scores obtained across individual observations in the held-out test subset.}
    \label{score_plot}
\end{figure}

Figure~\ref{score_plot} provides an additional visualization of the cosine-similarity scores across individual test observations.

The observed mean similarity indicates that the predicted and expert-reported representations were, on average, substantially aligned within the adopted representation space. However, cosine similarity quantifies agreement between numerical representations and is not equivalent to classification accuracy, precision, recall, or confirmation that the model independently identified the physical or operational cause of an occurrence.

\subsection{Qualitative Examination of Predicted Causal Representations}
\label{sec:qualitative_results}

\begin{figure}[H]
\centering

\begin{minipage}{0.47\textwidth}
    \centering
    \includegraphics[width=\textwidth]{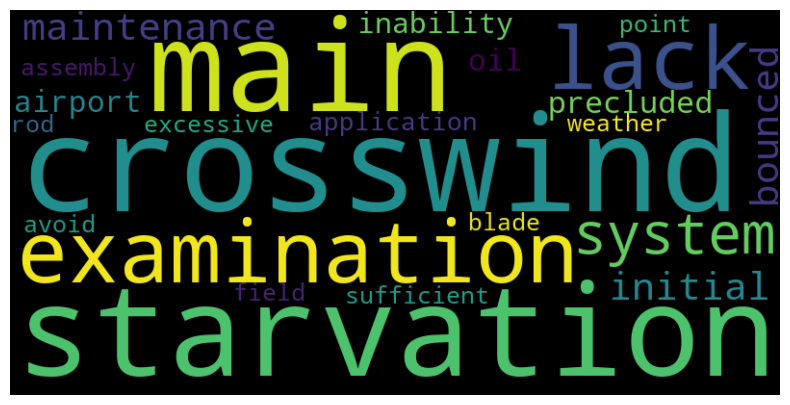}\\
    \textbf{(a)}
\end{minipage}
\hfill
\begin{minipage}{0.47\textwidth}
    \centering
    \includegraphics[width=\textwidth]{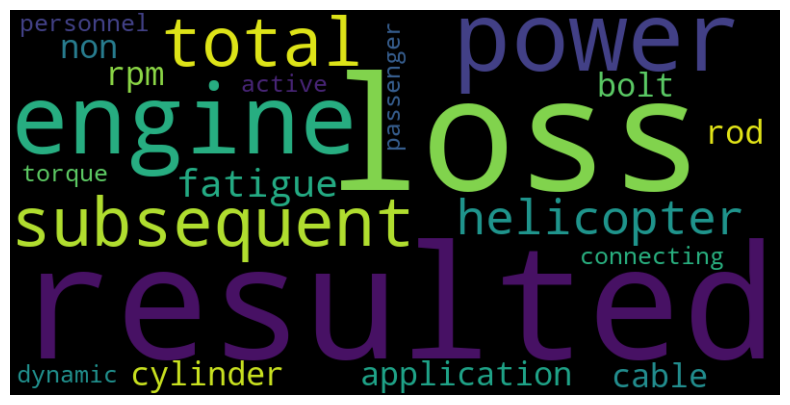}\\
    \textbf{(b)}
\end{minipage}

\vspace{0.5cm}

\begin{minipage}{0.47\textwidth}
    \centering
    \includegraphics[width=\textwidth]{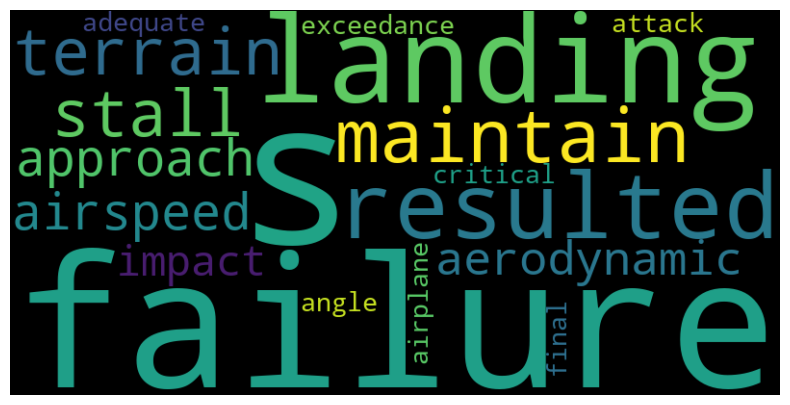}\\
    \textbf{(c)}
\end{minipage}
\hfill
\begin{minipage}{0.47\textwidth}
    \centering
    \includegraphics[width=\textwidth]{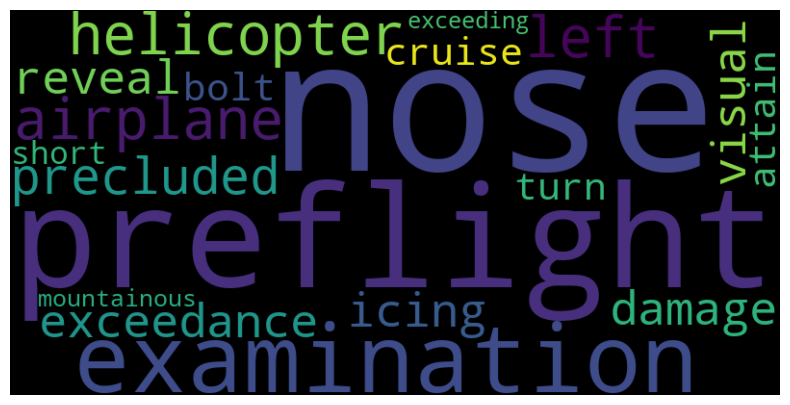}\\
    \textbf{(d)}
\end{minipage}

\caption{Word-cloud visualizations of terms recovered from predicted causal representations for four selected test observations. The examples contain terms associated with (a) operational, maintenance, and environmental factors; (b) engine and mechanical components; (c) flight dynamics and landing-related conditions; and (d) inspection, aircraft-component, and damage-related information. The word clouds are provided as qualitative illustrations of the information represented by the model outputs rather than as quantitative measures of causal importance.}
\label{word_cloud}
\end{figure}

To complement the aggregate similarity analysis, predicted representations from selected test observations were mapped back to the textual representation space and examined qualitatively. Figure~\ref{word_cloud}(a--d) presents word-cloud visualizations generated from four predicted outputs.

The examples demonstrate that the predicted representations contain aviation-related terms associated with technical, operational, environmental, and aircraft-performance information. Because the model output consists of recovered terms rather than a structured causal statement, these visualizations are intended to illustrate the semantic content of the predicted representations rather than to establish the relative importance of individual causal factors.

\subsection{Detailed Case Study}
\label{sec:case_study}

A representative test observation was examined in greater detail to illustrate the complete inference process and the interpretability of the resulting model output.

\begin{figure}[H]
    \centering
    \includegraphics[width=\textwidth]{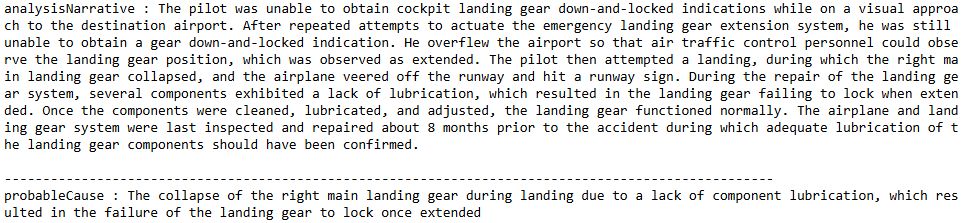}
    \caption{Example of the original analysis narrative and corresponding probable-cause information extracted from an NTSB investigation record before text preprocessing.}
    \label{investigation_report}
\end{figure}

Figure~\ref{investigation_report} presents the original analysis narrative and corresponding probable-cause information extracted from the NTSB investigation record before text preprocessing.

\begin{figure}[H]
    \centering
    \includegraphics[width=\textwidth]{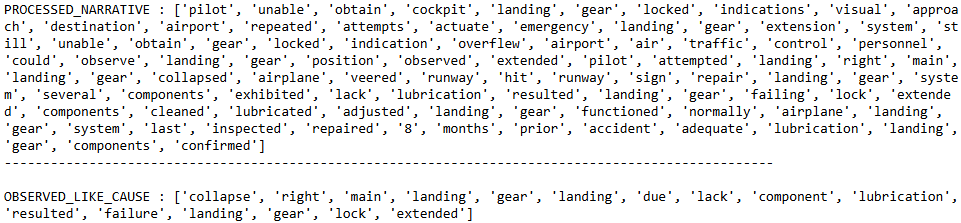}
    \caption{Processed textual representations of the analysis narrative and probable cause for the example investigation record after tokenization and removal of stop words and punctuation.}
    \label{investigation_report_processed}
\end{figure}

The narrative and probable-cause text were subsequently subjected to the preprocessing procedure described in Section~\ref{sec:preprocessing}. Figure~\ref{investigation_report_processed} shows the resulting word sequences after preprocessing.

\begin{figure}[H]
    \centering
    \includegraphics[width=\textwidth]{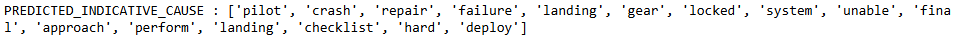}
    \caption{Terms recovered from the model-predicted causal representation for the example investigation record.}
    \label{predicted_indicative_cause}
\end{figure}

The processed narrative was converted to its numerical representation and supplied to the trained model. The decoder produced a predicted numerical causal representation, which was subsequently mapped back to the textual representation space. The recovered terms for this example are shown in Figure~\ref{predicted_indicative_cause}.

The recovered output contained terms including ``gear'', ``locked'', ``system'', ``unable'', ``deploy'', and ``checklist''. These terms are associated with the landing-gear-related circumstances represented in the corresponding investigation record. In particular, the combination of terms related to the landing gear, deployment, system state, and checklist procedures indicates that the learned representation captured information relevant to the landing-gear problem described in the occurrence.

Comparison with the investigator-reported probable cause shown in Figure~\ref{investigation_report} indicates semantic correspondence between the predicted terms and the causal information documented in the investigation record. Nevertheless, the recovered terms do not constitute a complete causal explanation and do not independently establish the sequence or mechanism of the occurrence. Rather, the example demonstrates how the model output can provide interpretable causal cues that may assist a human analyst in identifying information relevant to an occurrence.

\subsection{Performance Variation}
\label{sec:error_analysis}

Although the mean test-set cosine similarity was 0.786, the standard deviation of 0.120 indicates variation in semantic agreement across individual observations. As illustrated in Figures~\ref{normal_plot} and~\ref{score_plot}, a relatively small number of test observations produced similarity values below 0.5.

These results demonstrate that the model did not represent all narrative--cause relationships equally well. The original experiment did not include a systematic manual categorization of the low-similarity observations. Consequently, the available experimental evidence does not establish whether lower similarity was associated with particular causal-factor types, narrative complexity, uncommon terminology, multiple interacting causes, or limitations of the textual representation.

A systematic error analysis of high-, intermediate-, and low-similarity observations would be required to characterize these failure modes more precisely.

\section{Discussion}
\label{sec:discussion}

\subsection{Interpretation of the Findings}
\label{sec:discussion_findings}

This study investigated whether a variational-autoencoder-based representation-learning framework could capture the relationship between aviation investigation narratives and the corresponding probable causes reported by expert investigators. On the held-out test subset, the final model achieved a mean cosine similarity of 0.786 (SD = 0.120) between the predicted and expert-reported probable-cause representations. The distribution of similarity scores was concentrated toward the higher end of the observed range, although lower-similarity cases were also present.

These results indicate that the proposed model learned a measurable semantic relationship between the numerical representations of aviation investigation narratives and their corresponding expert-reported probable causes. Unlike a conventional classification approach, the proposed method does not require each occurrence to be assigned to one of a predefined set of causal categories. Instead, the encoder maps the investigation narrative to a lower-dimensional latent representation, from which the decoder estimates a numerical representation associated with the corresponding probable cause.

The qualitative results provide additional insight into the information retained by the learned representation. In the detailed example presented in Section~\ref{sec:case_study}, the model output contained terms including ``gear'', ``locked'', ``system'', ``unable'', ``deploy'', and ``checklist''. These terms were associated with the landing-gear-related circumstances represented in the corresponding investigation record. Similarly, the word-cloud examples in Figure~\ref{word_cloud} demonstrate that predicted representations contained terms associated with technical, operational, environmental, and aircraft-performance information.

The model output should nevertheless be distinguished from an investigator-generated probable-cause statement. The decoder produces a numerical representation that is subsequently mapped back to textual terms. Consequently, the present framework identifies indicative causal information rather than generating a complete causal explanation containing the relationships among events, contributing factors, and supporting evidence.

\subsection{Interpretation of Semantic Similarity}
\label{sec:discussion_similarity}

Cosine similarity was incorporated into both model development and evaluation because the objective of the framework is to preserve semantic agreement between predicted and expert-reported causal representations. Unlike a distance measure that is strongly influenced by vector magnitude, cosine similarity measures the angular relationship between two numerical representations. This characteristic makes it suitable for comparing distributed textual representations in which directional alignment can convey semantic relatedness.

The mean test-set cosine similarity of 0.786 therefore indicates the degree of alignment between the predicted and expert-reported probable-cause representations within the adopted representation space. It should not, however, be interpreted as an accuracy of 78.6\%, a precision of 78.6\%, or a probability that the predicted causal information is correct. Cosine similarity measures representation-level agreement rather than the correctness of an accident-investigation conclusion.

This distinction is particularly important in aviation accident and incident investigation. An official probable-cause determination may depend on evidence beyond the textual analysis narrative used as input to the model, including flight-recorder data, component examinations, maintenance information, meteorological evidence, operational procedures, and other investigative material. The proposed framework therefore operates on a subset of the information available to an investigation authority.

Furthermore, the interpretation of cosine similarity depends on the quality of the underlying numerical text representation. Limitations in representing domain-specific aviation terminology can propagate into both model training and evaluation. Consequently, the reported similarity provides evidence that the model learned correspondence within the adopted representation space, but additional evaluation would be required to establish performance at the level of individual causal factors or complete causal explanations.

\subsection{Relationship to Existing AI Approaches}
\label{sec:discussion_existing}

The proposed framework differs from several existing formulations of NLP-based aviation safety analysis. Topic-modeling approaches identify recurring themes within collections of aviation reports \cite{kenneth2018}, while supervised deep-learning approaches typically assign narratives to predefined occurrence or causal categories \cite{dong2021identifying,goblet2015identifying}. Such formulations are valuable when the required output taxonomy is known in advance, but they address a different problem from learning the relationship between an investigation narrative and its corresponding free-text probable cause.

Recent research has also introduced more explicit causal-information extraction and reasoning approaches. Retrieval-augmented generation has been investigated for causal identification from aviation accident reports \cite{ren2025rag}, while LLM reasoning guided by aviation-safety frameworks has been used to infer unsafe acts and associated conditions \cite{liu2025hfacs}. Transformer-based sequence-labeling approaches have also been proposed to identify explicit cause--effect structures in aviation narratives \cite{chen2026arcem}, and hybrid knowledge-graph--LLM architectures have been investigated for reconstructing causal relationships from accident information \cite{lin2026ai4airsafe}.

The present approach addresses the problem from a different perspective. Rather than extracting explicit causal spans, selecting a predefined category, or generating causal explanations using a large language model, the proposed VAE learns a probabilistic latent mapping between paired investigation-narrative and probable-cause representations. The resulting decoder output is subsequently mapped back to textual terms to provide an interpretable indication of the causal information captured by the learned representation.

The proposed formulation uses an architecture based on one-dimensional convolutional operations and a 20-dimensional latent representation. Nevertheless, the present experiments do not provide a direct empirical comparison with Transformer-, retrieval-, or LLM-based approaches. Consequently, no conclusion can be drawn from the current results regarding the relative performance of these model families. Controlled baseline comparisons using identical training and evaluation data would be required for such an assessment.

\subsection{Limitations}
\label{sec:limitations}

Several limitations should be considered when interpreting the results of this study.

First, the experiments were conducted using finalized NTSB investigation records from 2005--2020. Reporting conventions, terminology, occurrence characteristics, and investigative practices may differ across investigation authorities and time periods. The reported performance therefore does not establish generalization to other aviation safety databases.

Second, the textual representation used in the present framework is based on pretrained static word embeddings. Static representations assign a fixed representation to a word and therefore have limited ability to represent changes in meaning arising from linguistic context. This limitation may be particularly relevant for specialized aviation terminology and for narratives in which causal meaning depends on interactions among multiple statements.

Third, the model output consists primarily of indicative causal terms rather than a complete structured causal explanation. Although the recovered terms can provide interpretable information, they do not explicitly describe causal relationships, temporal ordering, or the evidence supporting a particular inference.

Fourth, cosine similarity was used as the principal quantitative evaluation measure. Although appropriate for assessing alignment between numerical representations, cosine similarity does not directly measure the factual correctness of individual causal factors. Evaluation using additional semantic measures and expert assessment would provide a more comprehensive characterization of model performance.

Fifth, the observed standard deviation and the presence of lower-similarity observations demonstrate that performance varies across individual reports. The original experiment did not include a systematic manual categorization of these cases. Consequently, the factors responsible for lower-performing predictions cannot be established from the available results alone.

Sixth, the dataset was divided using a random data-partitioning procedure rather than a temporal split. Although the held-out test subset contains observations not used directly for model fitting, reports from similar time periods and operational contexts may occur in both subsets. Temporal evaluation, in which models are trained on earlier reports and evaluated on later reports, would provide stronger evidence regarding performance on future occurrences.

Finally, models trained using historical investigation records can reproduce characteristics of the underlying data, including inconsistencies in terminology, reporting practices, and expert judgments. Human oversight therefore remains necessary when applying learned representations to safety-critical analysis.

\subsection{Future Research}
\label{sec:future_work}

Several extensions could strengthen the proposed framework. First, contextual language representations should be evaluated against the static embeddings used in the present study. Transformer-based sentence or document representations could provide improved handling of context-dependent aviation terminology and long-range relationships within investigation narratives.

Second, the proposed VAE should be compared with appropriate baseline approaches under identical train--validation--test conditions. Relevant comparisons include retrieval-based methods, deterministic neural networks, recurrent and convolutional architectures, and Transformer-based representations. Such experiments should consider not only semantic performance but also model size, computational requirements, and inference time.

Third, systematic error analysis should examine representative high-, intermediate-, and low-similarity predictions. Manual examination of these cases could determine whether performance is affected by narrative length, uncommon terminology, multiple interacting causes, occurrence type, or other characteristics of the investigation reports.

Fourth, future work should investigate more structured and explainable outputs. Rather than producing only a collection of associated terms, an extended framework could identify candidate causal factors together with the specific sentences or evidence within the investigation narrative that support each inference. Such evidence-linked outputs would provide analysts with a clearer basis for evaluating model predictions.

Finally, external and temporal validation should be performed using reports from later time periods and, where available, other aviation investigation authorities. This would provide stronger evidence regarding the robustness of the learned representations to changes in reporting style, terminology, operational environment, and investigative practice.

\section{Conclusions}
\label{sec:conclusions}

This study developed and evaluated a variational-autoencoder-based framework for inferring semantic representations of causal information from unstructured aviation investigation narratives. Rather than formulating causal-factor inference as classification into a predefined set of categories, the proposed approach learns a probabilistic latent mapping between investigation narratives and the corresponding probable causes reported by expert investigators. A cosine-similarity-based semantic reconstruction component was incorporated into the training objective to encourage alignment between the predicted and expert-reported causal representations.

The framework was evaluated using 20,919 finalized NTSB aviation accident and incident investigation records from 2005--2020. Following hyperparameter and architecture evaluation, the final configuration used the Adam optimizer with a learning rate of 0.001, a batch size of 16, and a 20-dimensional latent representation. On the held-out test subset, the model achieved a mean cosine similarity of 0.786 (SD = 0.120) between the predicted and expert-reported probable-cause representations.

Qualitative examination further demonstrated that predicted numerical representations could be mapped back to interpretable terms associated with causal information in the corresponding investigation records. In the detailed example examined in this study, the predicted representation recovered terms including ``gear'', ``locked'', ``system'', ``unable'', ``deploy'', and ``checklist'', which were associated with the landing-gear-related circumstances documented in the investigation. Additional predicted outputs contained terms associated with technical, operational, environmental, and aircraft-performance information.

On the held-out test set, the predicted and expert-reported probable-cause representations achieved a mean cosine similarity of 0.786 (SD = 0.120), indicating substantial semantic alignment within the representation space used in this study. This metric should not be interpreted as classification accuracy or as evidence that the model independently determines the true cause of an aviation occurrence.

The findings demonstrate the potential of probabilistic latent representation learning for extracting causal information from aviation investigation narratives. The proposed framework should be regarded as an AI-assisted approach for identifying semantic information associated with expert-reported probable causes rather than as a replacement for formal accident investigation or expert causal determination. Future research should investigate contextual language representations, systematic error analysis, comparisons with alternative model architectures, and temporal or external validation to establish the generalizability of the approach.


\end{document}